\documentclass{article}

\PassOptionsToPackage{numbers, compress}{natbib}

\usepackage[main, final]{neurips_2026}

\usepackage[utf8]{inputenc} 
\usepackage[T1]{fontenc}    
\usepackage{hyperref}       
\usepackage{url}            
\usepackage{booktabs}       
\usepackage{amsfonts}       
\usepackage{nicefrac}       
\usepackage{microtype}      
\usepackage{xcolor}         
\usepackage{amsmath}
\usepackage{graphicx}
\usepackage{multirow}
\usepackage{wrapfig}

\usepackage{colortbl}
\definecolor{lightgray}{gray}{0.9}
\usepackage{array}
\definecolor{lightblue}{RGB}{91,155,213}

\usepackage{pifont}

\makeatletter
\renewcommand{\@trackname}{}
\makeatother

\title{Null-Space–Constrained Contrastive Visual Forgetting for MLLM Unlearning}
\author{%
  Yuhang Wang$^{1}$ \And
  Zhenxing Niu$^{1}$ \And
  Haoxuan Ji$^{2}$ \And
  Guangyu He$^{1}$ \And
  Linlin Zhang$^{1}$ \And
  Haichang Gao$^{1}$ \\
  $^{1}$ School of Computer Science and Technology, Xidian University \\
  $^{2}$ Xi'an Jiaotong University
}

\begin{document}

\maketitle

\vspace{-2em} 
\begin{abstract}
The core challenge of machine unlearning is to strike a balance between target knowledge removal and non-target knowledge retention. In the context of Multimodal Large Language Models (MLLMs), this challenge becomes even more pronounced, as knowledge is further divided into visual and textual modalities that are tightly intertwined. In this paper, we introduce an MLLM unlearning approach that aims to forget target visual knowledge while preserving non-target visual knowledge and all textual knowledge. Specifically, we freeze the LLM backbone and achieve unlearning by fine-tuning the visual module. 
First, we propose a \emph{Contrastive Visual Forgetting} (CVF) mechanism to separate target visual knowledge from retained visual knowledge, guiding the representations of target visual concepts toward appropriate regions in the feature space.
Second, we identify the \emph{null space} associated with retained knowledge and constrain the unlearning process within this space, thereby significantly mitigating degradation in knowledge retention. Third, beyond \emph{static} unlearning scenarios, we extend our approach to \emph{continual unlearning}, where forgetting requests arrive sequentially. Extensive experiments across diverse benchmarks demonstrate that our approach achieves a strong balance between effective forgetting and robust knowledge retention.
\end{abstract}

\vspace{-1em} 
\section{Introduction}
Multimodal Large Language Models (MLLMs)~\cite{wang2024qwen2,liu2023visual} have demonstrated strong multimodal capabilities and are increasingly deployed in real-world applications~\cite{yang2025pencil,fei2025path}. However, recent studies reveal that large models tend to memorize their training data, thereby introducing serious privacy leakage risks~\cite{huo2025mmunlearner,liu2024towards,dontsov2025clear}. Retraining a large model after excluding such privacy-sensitive data could address this issue, but its prohibitive computational cost makes it impractical, especially for the scenarios involving frequent or \emph{continual} forgetting requests. To address this challenge, machine unlearning techniques~\cite{li2023international,cao2015towards,gupta2021adaptive,sekhari2021remember,ullah2021machine} have recently attracted growing research attention, as they can precisely remove the influence of specific training data from an already trained large model. The core challenge in machine unlearning lies in balancing the trade-off between removing the target knowledge and preserving non-target knowledge.

While machine unlearning for LLMs has achieved notable progress~\cite{maini2024tofu,rafailov2023direct,zhang2024negative,wang2025gru}, unlearning for MLLMs remains in its nascent stage. When extending from LLMs to MLLMs, the knowledge stored in the model is further divided into visual and textual modalities. For a target entity (\emph{e.g.}, a person of interest), an MLLM encodes two types of knowledge: \emph{textual knowledge}, which captures factual information (\emph{e.g.}, a person’s home address), and \emph{visual knowledge}, which represents visual characteristics (\emph{e.g.}, a person’s facial appearance). Unlike text-only LLM unlearning—where the goal is to remove target textual knowledge—MLLM unlearning is typically defined as forgetting target visual knowledge while preserving non-target visual knowledge and all textual knowledge (including both target and non-target entities' textual knowledge)~\cite{huo2025mmunlearner,wang2025mllm}. This makes MLLM unlearning significantly more challenging than LLM unlearning, as target visual knowledge may be entangled not only with non-target visual knowledge but also with target textual knowledge.

To address this challenge, we propose an MLLM unlearning framework that improves the \emph{forgetting-retention trade-off} and facilitates the selective separation of \emph{target visual knowledge from preserved multimodal knowledge.}
Specifically, since the textual knowledge of both target and non-target entities should be preserved, we freeze the LLM backbone and fine-tune only the visual module (\emph{i.e.}, the ViT encoder and the projector). This design restricts the update scope to the part of the model most directly responsible for visual understanding, while reducing unnecessary degradation to textual knowledge. However, this design choice also makes the forgetting–retention trade-off more challenging: the visual module contains significantly fewer trainable parameters than the LLM backbone, rendering the task of forgetting target visual knowledge without compromising retained visual knowledge a more delicate optimization problem.

To this end, we propose a \emph{Contrastive Visual Forgetting} (CVF) mechanism to facilitate the forgetting–retention trade-off. 
Most existing unlearning methods rely on output-level supervision, where the objective is imposed on the final model outputs. While this is effective when the entire MLLM is updated, it is less suitable in our setting, where only the visual module is optimized. In MLLMs, the visual module is followed by the LLM backbone, meaning that output-level supervision must propagate through the LLM before reaching the visual module, thereby weakening the effectiveness of supervision.
To address this limitation, we directly inject supervision into the visual module (\emph{i.e.}, intermediate-level supervision) and introduce the CVF mechanism to forget target visual knowledge while preserving non-target visual knowledge. Specifically, we leverage features from a \emph{reference} model as intermediate-level supervision and guide the representations of target visual concepts toward appropriate regions in the feature space. Importantly, these regions are defined by anchor representations associated with retained knowledge, thereby preventing harmful drift that could degrade knowledge retention.

Furthermore, inspired by null-space learning methods for LLM knowledge editing~\cite{fang2024alphaedit}, we propose a \emph{null-space–constrained unlearning} (NCU) scheme to further mitigate interference between the forgetting and retention processes. The key idea is to identify subspaces that are weakly associated with retained knowledge and constrain the unlearning updates to lie within this subspace. NCU integrates seamlessly with CVF and introduces an additional structural constraint on the optimization process, resulting in a more balanced trade-off between effective forgetting and robust knowledge preservation.

We conduct extensive experiments on three multimodal unlearning benchmarks, including MLLMU, UMU, and CLEAR. The results show that our method achieves a favorable balance between effective target forgetting, retained knowledge preservation, and general multimodal utility. In addition, beyond \emph{static} unlearning, we study \emph{continual} unlearning, where forgetting requests arrive sequentially over time. This setting is practically important but remains underexplored in prior work. Our method can be naturally extended to continual unlearning and consistently outperforms prior baselines in this more challenging scenario.
\vspace{-1em}

\section{Preliminary}
\vspace{-1em} 
An MLLM $\mathcal{M}$ typically consists of three components: a vision encoder $\mathcal{V}$, a large language model $\mathcal{W}$, and a linear projector $\mathcal{P}$ that bridges the two.
Generally, $\mathcal{M}$ is trained on a visual instruction dataset $\mathcal{D}=\{(v_i,q_i,a_i)\}_{i=1}^{N}$, where $v_i$ denotes the input image, $q_i$ the textual instruction/question, and $a_i$ the corresponding answer.
Given a sample $(v_i,q_i,a_i)$, the image $v_i$ is first processed by the vision encoder and the projector, yielding the visual embedding $v_i^{e}=\mathcal{P}(\mathcal{V}(v_i))$.
The instruction $q_i$ is converted into a textual embedding $q_i^{e}$, which is then concatenated with $v_i^{e}$ and fed into the language model $\mathcal{W}(q^e_i \oplus v^e_i)$. Finally, the MLLM generates an output that is expected to align with the target answer $a_i$.

Following standard unlearning practice, the dataset $\mathcal{D}$ is partitioned into a forget set $\mathcal{D}_f$ that contains samples associated with the \emph{target entity}, and a retain set $\mathcal{D}_r$ that contains all other samples. In the context of MLLM unlearning, we distinguish two types of knowledge associated with the target entity: visual knowledge and textual knowledge. We define MLLM unlearning for a target entity as the process of \emph{erasing its visual knowledge while preserving its textual knowledge; meanwhile, both the visual and textual knowledge associated with other entities should remain unaffected}.

To evaluate the performance of MLLM unlearning, we employ both a visual question answering (VQA) task and a text-only question answering (QA) task to probe the visual and textual knowledge retained by the model.
Specifically, for a target-entity sample $(v, q, a) \in \mathcal{D}_f$, we query the model both with the image $v$ and without the image, respectively.
\emph{We regard MLLM unlearning as successful if the model fails the VQA task while still succeeding in the corresponding QA task}. Meanwhile, for samples $(v,q,a)\in\mathcal{D}_r$ that are unrelated to the target entity, the unlearned model is expected to successfully perform both the VQA and QA tasks.
\vspace{-1em}

\section{Our Approach}
\vspace{-1em} 
Machine unlearning aims to erase the target knowledge while preserving the remaining knowledge. Knowledge is generally regarded as being stored in a model’s parameters~\cite{li2024cmmlu,du2024evaluating}. In the context of LLM unlearning, since it is unclear which specific parameters encode the knowledge to be forgotten, unlearning is typically achieved by fine-tuning the \emph{entire} model using \emph{output-level} supervision signals. For example, given a sample $(q_i,a_i) \in \mathcal{D}_f$, gradient ascent (GA)~\cite{thudi2022unrolling} is applied to encourage the model output $\mathcal{W}(q_i)$ to deviate as far as possible from the ground-truth answer $a_i$.


In contrast, for the MLLM unlearning problem, the objective is to forget the target entity’s visual knowledge while retaining all textual knowledge. Accordingly, we propose to fine-tune only the visual module of the MLLM. On the one hand, freezing the LLM backbone can largely preserve textual knowledge, including both target and non-target entities’ textual information. On the other hand, our design is further motivated by recent studies on the internal mechanisms of MLLMs~\cite{huang2024commonsense,yu2024understanding}. These works reveal that the visual inference process in MLLMs typically involves two stages: the first stage visually recognizes the entity (termed \emph{identification}), and the second stage associates the recognized entity with its stored factual knowledge (termed \emph{extraction}). Empirical evidence suggests that the visual module is primarily responsible for the identification, whereas the LLM module mainly performs the extraction. Therefore, by disrupting the identification process through unlearning in the visual module, we can effectively prevent the MLLM from producing the target entity’s visual knowledge.

However, this design introduces a new challenge: 
existing unlearning methods typically adopt an \emph{output-level} supervision scheme, where the unlearning objective is defined on the final outputs of the MLLM. While this strategy is well suited for fine-tuning the entire MLLM, it is not appropriate for our setting, where only the visual module is fine-tuned. This is because an MLLM consists of a visual module followed by an LLM backbone, and output-level supervision signals are significantly attenuated by the time they are back propagated to the visual module.
To address this limitation, we inject the supervision signal directly into the visual module (\emph{intermediate-level} supervision). Furthermore, we develop two novel schemes to improve the performance of unlearning, \emph{i.e.,} Contrastive Visual Forgetting (CVF) and Null-space-Constrained Unlearning (NCU), which will be described in detail in the following sections.

\vspace{-0.8em}
\subsection{Contrastive Visual Forgetting}
\vspace{-0.8em}

\begin{wrapfigure}{r}{0.51\columnwidth}
    \vspace{-0.8em}
    \centering
    \includegraphics[width=0.5\columnwidth]{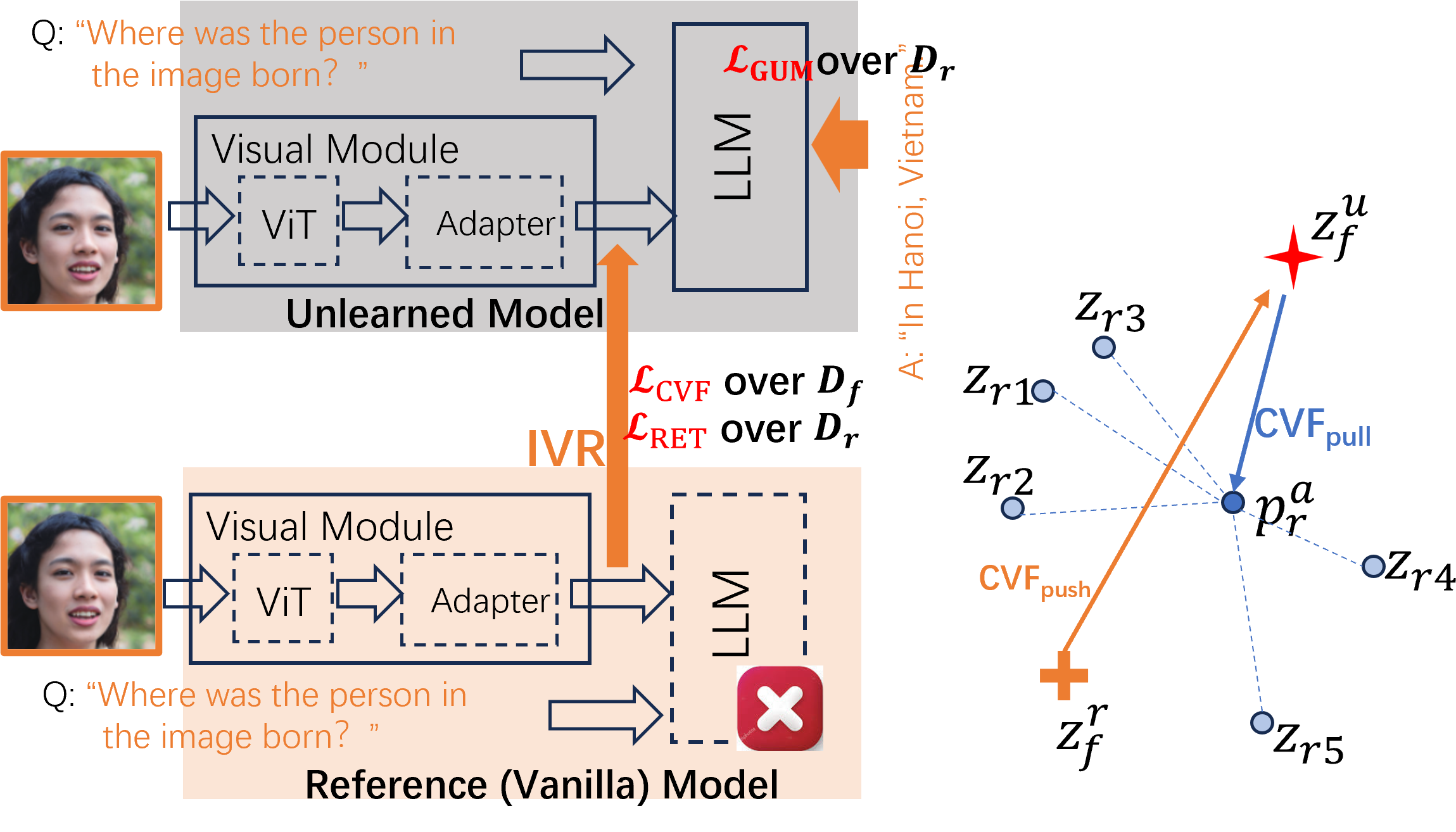}
    \caption{Overview of our CVF mechanism in MLLM unlearning.}
    \label{fig1}
    \vspace{-1.2em}
\end{wrapfigure}
In our approach, we use the original model as a \emph{reference} model and take its Intermediate Visual Representations (IVRs) as supervision signals. As illustrated in Fig.~\ref{fig1}, the IVRs correspond to the outputs of the visual module, which also serve as the inputs to the LLM backbone. To forget a target visual knowledge, we encourage the IVRs produced by the unlearned model to deviate from those of the reference model. Conversely, to retain non-target visual knowledge, we enforce the IVRs of the unlearned model to remain close to their counterparts in the reference model. 
Another advantage of intermediate-level supervision is that it eliminates the need to load the full reference model into memory; instead, only the visual module is required to obtain the supervision signal.

Specifically, we introduce a \emph{Contrastive Visual Forgetting} (CVF) mechanism, which adapts contrastive learning to explicitly forget target visual knowledge. For a forgetting sample $(v,q,a)\in\mathcal{D}_f$, the corresponding IVRs are computed as follows. Let $\mathcal{E}$ denote the visual module, which consists of a vision encoder $\mathcal{V}$ and a projector $\mathcal{P}$. We use $\mathcal{E}_\ell(\cdot)$ to denote the visual representation at level $\ell$. The token-level IVRs of the \emph{u}nlearned model and the \emph{r}eference model are obtained as follows:
\begin{equation}
H_f^{u} = \mathcal{E}_{\ell}(v;\theta_0,\phi),
\qquad
H_f^{r} = \mathcal{E}_{\ell}(v;\theta_0),
\end{equation}
where $\theta_0$ denotes the parameters of the reference model, and $\phi$ denotes the LoRA parameters of the unlearned model. 
We then apply a pooling operator $g(\cdot)$ (specifically, mean pooling over tokens) to aggregate the token-level IVRs and obtain the global IVRs:
\begin{equation}
z_f^{u} = g(H_f^{u}), \qquad z_f^{r} = g(H_f^{r}).
\end{equation}

Meanwhile, we maintain a negative queue $\mathcal{Q}$ that stores reference IVRs from previous iterative steps.
We normalize them and compute cosine similarities with temperature $\tau$:
\begin{equation}
\bar{z}=\frac{z}{\lVert z\rVert_2},
\qquad
\mathrm{sim}(u,r)=\frac{u^\top r}{\tau}.
\end{equation}

We propose an InfoNCE-like loss function and minimize the probability of matching between the reference and unlearned IVRs as:
\begin{equation}
\mathcal{L}_{\mathrm{CVF}\text{-}{push}}
=
-\log
\frac{\sum\limits_{k\in\mathcal{Q}} \exp\!\big(\mathrm{sim}(\bar{z}_f^{u},\bar{k})\big)}
{\exp\!\big(\mathrm{sim}(\bar{z}_f^{u},\bar{z}_f^{r})\big)
+\sum\limits_{k\in\mathcal{Q}} \exp\!\big(\mathrm{sim}(\bar{z}_f^{u},\bar{k})\big)}. \label{eq:push}
\end{equation}

Unlike the InfoNCE loss~\cite{oord2018representation}, whose numerator consists of positive samples, our formulation places negative samples in the numerator. This design explicitly reflects our objective: to push the unlearned IVRs $z_f^{u}$ away from the reference IVRs $z_f^{r}$, rather than pulling them closer.

Note that pushing the unlearned IVRs away from the reference IVRs can be trivially achieved by minimizing the negative mean squared error as follows,
\begin{equation}
\mathcal{L}_{\mathrm{nMSE}}=-\left\lVert z_f^{u} - z_f^{r}\right\rVert^2.
\end{equation}

However, our experimental results demonstrate that the proposed CVF loss achieves a substantially better balance between effective forgetting and knowledge retention than this naive alternative.

Furthermore, our empirical studies show that naively pushing the unlearned IVRs arbitrarily far from the reference IVRs, without a clear target direction, can severely harm the preservation of remaining knowledge. To this end, we enhance our CVF mechanism by providing a principled destination toward which the unlearned IVRs are pushed. 

Concretely, we introduce an \emph{anchor} IVRs $p^a_r\in R^{d}$ associated with retained knowledge.
At each iteration, $p^a_r$ is computed as the mean of the reference IVRs over retaining samples within the mini-batch $(v,q,a)\in\mathcal{B}_r$. We then encourage the unlearned IVRs to remain close to this anchor by minimizing the following objective:
\begin{equation}
\mathcal{L}_{\mathrm{CVF}\text{-}pull}
=
1-\frac{\langle \bar{z}_f^{u}, \bar{p}^a_r\rangle}{\lVert \bar{z}_f^{u}\rVert_2\lVert \bar{p}^a_r\rVert_2}. \label{eq:pull}
\end{equation}

Finally, the final CVF objective is
\begin{equation}
\mathcal{L}_{\mathrm{CVF}}
=
\lambda\,\mathcal{L}_{\mathrm{CVF}\text{-}push}
+
\,\mathcal{L}_{\mathrm{CVF}\text{-}pull},
\end{equation}
where $\lambda$ controls the trade-off between pushing the unlearned IVRs away from the reference IVRs and pulling them toward the anchor IVRs, respectively.

\paragraph{Retention of Non-target Visual Knowledge.}
Our CVF mechanism primarily focuses on effectively forgetting target visual knowledge, which may inadvertently impair the retention of non-target visual knowledge. Therefore, we explicitly encourage the preservation of non-target visual knowledge by additionally leveraging IVRs from the reference model as supervision signals.

Specifically, for each sample $(v,q,a)\in\mathcal{D}_r$, we first obtain the token-level IVRs from both the unlearned model and the reference model as follows:
\begin{equation}
H_r^{u} = \mathcal{E}_{\ell}(v;\theta_0,\phi),
\qquad
H_r^{r} = \mathcal{E}_{\ell}(v;\theta_0),
\end{equation}

We then define the retention loss term as a token-wise mean squared error:
\begin{equation}
\mathcal{L}_{\mathrm{RET}}
=
\frac{1}{N}\left\lVert H_r^{u}-H_r^{r}\right\rVert_F^2, \label{eq:ret}
\end{equation}
where $\lVert\cdot\rVert_F$ is the Frobenius norm.
By explicitly encouraging the IVRs of unlearning model to match that of reference model on $\mathcal{D}_r$, our $\mathcal{L}_{\mathrm{RET}}$ can retain non-target visual knowledge.

\textbf{General Utility Maintenance.}
Although our intermediate-level supervision scheme can effectively balance the forgetting–retention trade-off, our empirical results further demonstrate that incorporating output-level supervision can bring additional improvements. Moreover, the \emph{General Utility} for an unlearned MLLM is a crucial metric for evaluating unlearning performance. This utility is typically assessed on real-world VQA datasets that are distinct from both the target and non-target VQA datasets. We find that output-level supervision can also benefit the maintenance of an MLLM’s general utility.

Specifically, for each sample $(v,q,a)\in\mathcal{D}_r$, when feeding the image $v$ and query $q$ into the MLLM, we maximize the likelihood of generating the ground-truth answer $a$:
\begin{equation}
\mathcal{L}_{\text{GUM}}
=
-\log p_{\theta}\!\left(a \mid v, q\right), \label{eq:gum}
\end{equation}
where $p_{\theta}$ denotes the likelihood function of the unlearned model.

In summary, our final loss function combines all three aforementioned loss terms over both the forget set $\mathcal{D}_f$ and the retain set $\mathcal{D}_r$:
\begin{equation}
\min_{\theta}\;
\alpha\,
\mathcal{L}_{\text{CVF}}
+
\beta\,
\mathcal{L}_{\text{RET}}
+
\mathcal{L}_{\text{GUM}},
\end{equation}
where $\alpha$ and $\beta$ are weighting coefficients that balance the three loss terms.

\vspace{-0.8em}
\subsection{Null-space–Constrained Unlearning}
\vspace{-0.8em}
Our CVF mechanism improves the separation between target visual knowledge and non-target visual knowledge through a carefully designed objective function; however, the forgetting and retained knowledge are not \emph{explicitly} decoupled during unlearning.
In this section, inspired by the null-space learning~\cite{fang2024alphaedit} developed for LLM knowledge editing, we propose a \emph{Null-space Constrained Unlearning} (NCU) method to explicitly decouple the knowledge forgetting from the knowledge retention.

In the context of LLM knowledge editing, particularly \emph{lifelong editing}, editing requests typically arrive sequentially and must be handled one by one. A critical challenge is to ensure that a newly applied editing operation does not interfere with or overwrite the effects of previous edits. To address this issue, prior to performing the current editing operation, one can identify the null space associated with all previous editing operations and constrain the new edit to lie within this null space. As a result, the current editing operation can be carried out without affecting any of the previously applied edits.

In the context of unlearning task, if we can identify the null space corresponding to the knowledge that should be retained and restrict the forgetting operation in this null space, we can effectively erase the target knowledge while largely preserving the retained knowledge. In the following, we describe our NCU approach in detail.

\paragraph{Identify subspace associated with retained knowledge.}
Our unlearning procedure is realized by fine-tuning the visual module of the MLLM. Let $\mathcal{L}$ denote the set of all layers within this visual module. We first collect layer-wise activations on the retain set $\mathcal{D}_r$ using the reference model. For each sample $(v,q,a)\in\mathcal{D}_r$, let $x_{\ell}\in R^d$ denote the activations at layer $\ell$ corresponding to a single token.
By aggregating the activations of all tokens, we construct an activation matrix
\begin{equation}
X_{\ell} = [x_{\ell}^{(1)}, x_{\ell}^{(2)}, \dots, x_{\ell}^{(n_{\ell})}]^{\top}\in R^{n_{\ell}\times d},
\end{equation}
where $n_{\ell}$ denotes the number of tokens at layer $\ell$.
We then compute the covariance matrix
\begin{equation}
C_{\ell} = \frac{1}{n_{\ell}} X_{\ell}^{\top}X_{\ell}\in R^{d\times d}.
\end{equation}
Intuitively, $C_{\ell}$ captures the principal directions that encode the retained knowledge at layer $\ell$. We then apply a Singular Value Decomposition (SVD) to $C_{\ell}$:
\begin{equation}
C_{\ell} = U_{\ell}\Lambda_{\ell}U_{\ell}^{\top},
\end{equation}
where $\Lambda_{\ell}=\mathrm{diag}(\lambda_{\ell,1},\dots,\lambda_{\ell,d})$ with $\lambda_{\ell,1}\le\dots\le\lambda_{\ell,d}$, and columns of $U_{\ell}$ are the corresponding eigenvectors.

Small eigenvalues correspond to directions that are rarely activated by retained knowledge.
Accordingly, we define the null space associated with retained knowledge as the subspace spanned by the eigenvectors corresponding to the $r$ smallest eigenvalues:
\begin{equation}
U_{\ell}^{\perp} = U_{\ell}[:, 1:r]\in R^{d\times r}.
\end{equation}

\paragraph{Null-space–constrained unlearning.}
Our MLLM unlearning approach achieves knowledge forgetting by fine-tuning the visual module, where the fine-tuning is implemented via a Low-Rank Adaptation (LoRA) scheme~\cite{hu2022lora}. Let $W$ denote the original model weights. After unlearning, the updated weights are given by $W+ \Delta W =W+ BA$, where $\Delta W =BA$ represents the LoRA parameters learned during fine-tuning. For each layer $\ell$ in the visual module, we denote the layer-wise weight by $W_{\ell}$ and the corresponding LoRA parameters by $\Delta W_{\ell} =A_{\ell}B_{\ell}$, where both $A_{\ell}$ and $B_{\ell}$ are low-rank matrices with rank $k$:
\begin{equation}
\Delta W_{\ell} = B_{\ell}A_{\ell}, \quad
A_{\ell}\in R^{r\times d_{\text{in}}},\;
B_{\ell}\in R^{d_{\text{out}}\times r}.
\end{equation}
During our null-space–constrained unlearning, we initialize 
\begin{equation}
A_{\ell} = (U_{\ell}^{\perp})^{\top},
\end{equation}
and freeze $A_{\ell}$ throughout unlearning, while optimizing only $B_{\ell}$. Our NCU differs fundamentally from previous unlearning methods, which randomly initialize both $A_{\ell}$ and $B_{\ell}$ and allow them to be jointly updated during fine-tuning.

Notably, our NCU scheme can be naturally extended to address the continual unlearning problem (forgetting requests arrive sequentially).
Specifically, we freeze $A_{\ell}$ throughout the continual unlearning procedure and use the optimized $B_{\ell}$ from the previous forgetting task as the initialization for the unlearning optimization of the subsequent task.

Let us explain why our NCU scheme can effectively decouple knowledge forgetting from knowledge retention.
We first review the definition of the \emph{null space}~\cite{fang2024alphaedit}.
Specifically, given two matrices $A$ and $B$, $B$ is in the null space of $A$ if and only if $BA = 0$. 
For the unlearned model, we have
\begin{equation}
y_{\ell} = (W_{\ell} + \Delta W_{\ell})x_{\ell}
= W_{\ell}x_{\ell} + B_{\ell}A_{\ell}x_{\ell}.
\end{equation}
Thus, for activations $x_{\ell}$ associated with retained knowledge, we have $A_{\ell}x_{\ell}\approx 0$, and consequently $B_{\ell}A_{\ell}x_{\ell}\approx 0$. This implies that the LoRA update contributes negligibly to the layer output on $\mathcal{D}_r$, thereby exerting minimal impact on the retained knowledge.

\vspace{-0.8em}
\subsection{Integration of NCU and CVF}
Our NCU scheme can be seamlessly integrated with the proposed CVF mechanism. While CVF focuses on designing effective objective functions, NCU imposes structural constraints on the optimization process. In practice, NCU is applied once as an offline preprocessing step on the retain set $\mathcal{D}_r$ to initialize the LoRA parameters $A_{\ell}$. Subsequently, LoRA fine-tuning performs unlearning by optimizing the CVF objectives solely with respect to the parameters $B_{\ell}$.

\begin{table*}[t]
\centering
\scriptsize  
\setlength{\tabcolsep}{1pt} 
\renewcommand{\arraystretch}{0.9} 

\begin{tabular}{l|cccccc|cccccc}
\toprule
\multirow{1}{*}{\textbf{Methods}} &
\multicolumn{6}{c|}{\textbf{LLaVA-1.5-7B}} &
\multicolumn{6}{c}{\textbf{Qwen2\text{-}VL\text{-}7B}} \\
\midrule
\multicolumn{13}{c}{\textbf{UMU\text{-}Bench}} \\
\midrule
& Forget VQA & Forget QA & Retain VQA & Retain QA & RW VQA & RW QA
& Forget VQA & Forget QA & Retain VQA & Retain QA & RW VQA & RW QA \\
& Acc($\downarrow$) & Acc($\uparrow$) & Acc($\uparrow$) & Acc($\uparrow$) & Acc($\uparrow$) & Acc($\uparrow$)
& Acc($\downarrow$) & R\text{-}L($\uparrow$) & Acc($\uparrow$) & R\text{-}L($\uparrow$) & Acc($\uparrow$) & Acc($\uparrow$) \\
\midrule
Vanilla      & 76.0 & 82.0 & 76.3 & 69.5 & 44.8 & 49.0 & 62.0 & 70.1 & 71.8 & 60.5 & 48.0 & 45.1 \\
GA           & 53.7 & 58.0 & 59.1 & 66.4 & 40.5 & 48.0 & 55.4 & 53.7 & 55.6 & 54.8 & 31.2 & 32.6 \\
GA\_Diff     & 54.0 &  60.0 & 56.6 & 65.1 & 40.7 & 45.7 & 50.0 & 52.0 & 56.3 & 58.1 & 30.1 & 30.3 \\
KL\_Min      & 50.0 & 56.0 & 58.1 &  67.6 & 39.7 & 47.3 & 52.0 & 52.0 & 57.0 & 56.5 & 29.7 & 30.1 \\
NPO         & 54.0 & 59.0 &  65.1 & 62.7 &  42.1 & 41.0 & 46.7 & 48.0 & 52.3 & 56.0 & 34.2 & 31.7 \\         
MANU         & 50.0 & 56.0 & 53.8 & 60.0 & 41.1 & 45.1 & 42.0 & 47.4 & 56.3 & 55.4 & 33.5 & 34.6 \\
MMU  &  46.0 & 52.2 & 54.4 & 59.2 & 40.1 &  48.3 & 41.8 & 46.7 & 56.6 & 57.3 & 34.5 & 39.1 \\
\rowcolor{lightgray}
\textbf{Ours}          & \textbf{44.2} & { \textbf{82.0}} & \textbf{74.4} & { \textbf{69.5}} & \textbf{43.2} & { \textbf{49.0}} & \textbf{40.4} & { \textbf{70.1}} & \textbf{63.3} & { \textbf{60.5}} & \textbf{36.1} & { \textbf{45.1}} \\

\midrule
\multicolumn{13}{c}{\textbf{CLEAR}} \\
\midrule
& Forget VQA & Forget QA & Retain VQA & Retain QA & RW VQA & RW QA
& Forget VQA & Forget QA & Retain VQA & Retain QA & RW VQA & RW QA \\
& Acc($\downarrow$) & R\text{-}L($\uparrow$) & Acc($\uparrow$) & R\text{-}L($\uparrow$) & Acc($\uparrow$) & Acc($\uparrow$) 
& Acc($\downarrow$) & R\text{-}L($\uparrow$) & Acc($\uparrow$) & R\text{-}L($\uparrow$) & Acc($\uparrow$) & Acc($\uparrow$) \\
\midrule
Vanilla      & 63.3 & 0.367 & 54.0 & 0.352 & 53.7 & 85.4 & 67.0 & 0.116 & 70.9 & 0.098 & 69.2 & 91.4 \\
GA           & 57.4 & 0.153 & 48.4 & 0.176 & 51.8 & 83.4 & 55.3 &  0.123 & 62.4 & 0.083 & 65.9 & 86.8 \\
GA\_Diff     & 47.3 & 0.197 & 43.4 & 0.220 & 47.7 & 73.5 & 63.3 & \textbf{0.125} &  67.4 & 0.088 & 68.4 &  92.7 \\
KL\_Min      & 40.4 & 0.270 & 38.1 & 0.274 & 51.5 & 82.8 & 67.0 & 0.120 & 68.9 &  0.098 & 68.4 & 90.7 \\
NPO          & 40.4 & 0.285 & 38.6 & 0.282 & 52.9 & 83.4 & 62.8 & 0.103 & 68.3 & 0.091 &  68.9 & 88.7 \\
MANU         & 39.8  & 0.263 & 25.8  & 0.224 & 49.8  & 68.3  & 61.3  & 0.107 & 65.2  & 0.094 & 66.2  & 87.3  \\
MMU         &  36.2 &  0.348 & 46.6 &  0.338 &  52.3 &  84.1  & 50.0  & 0.123 & 68.9  & 0.094 & 68.9  & 87.3  \\
\rowcolor{lightgray}
\textbf{Ours}          & \textbf{32.4} & { \textbf{0.367}} & \textbf{50.0} & { \textbf{0.352}} & \textbf{53.0} & { \textbf{85.4}} & \textbf{47.6} & { \textbf{0.116}} & \textbf{70.9} & { \textbf{0.098}} & \textbf{69.0} & { \textbf{91.4}} \\
\midrule
\multicolumn{13}{c}{\textbf{MLLMU\text{-}Bench}} \\
\midrule
& Forget VQA & Forget QA & Retain VQA & Retain QA & RW VQA & RW QA
& Forget VQA & Forget QA & Retain VQA & Retain QA & RW VQA & RW QA \\
& Acc($\downarrow$) & Acc($\uparrow$) & Acc($\uparrow$) & Acc($\uparrow$) & Acc($\uparrow$) & Acc($\uparrow$)
& Acc($\downarrow$) & Acc($\uparrow$) & Acc($\uparrow$) & Acc($\uparrow$) & Acc($\uparrow$) & Acc($\uparrow$)
\\
\midrule
Vanilla      & 45.8 & 38.4 & 45.2 & 37.5 & 47.4 & 54.9 & 55.2 & 55.0 & 56.0 & 58.6 & 77.3 & 77.5 \\
GA           & 43.2 & 32.5 & 45.0 & 32.2 &  47.0 & 54.7 & 50.4 & 46.7 & 51.5 & 57.6 & 74.4 & 77.8 \\
GA\_Diff     & 40.0 & 33.6 & 44.3 & 31.5 & 46.6 & 53.6 & 54.4 & 52.8 & 38.8 & 54.4 & 74.5 & 77.0 \\
KL\_Min      & 42.4 & 33.6 & 44.9 & 32.0 & \textbf{47.4} & 54.6 & 45.6 & 45.3 & 35.9 & 55.6 & 74.8 & 77.1 \\
NPO          & 43.2 & 33.6 &  44.9 & 32.2 &  47.0 & 54.7 & 49.6 & 50.4 & 49.5 & 53.3 & 75.2 & 78.3 \\
MANU         & 41.6  & 33.6  & 43.2  & 32.1  & 46.3  &  54.8  & 43.2  & 47.2 & 43.2  & 49.4 & 73.7  & 76.2  \\
MMU         &  31.2  & 34.2  & 44.2  & 35.1  & 46.7  & 54.9  & 44.0  & 54.4 & 47.4  & 55.7 & 75.2  & 77.3  \\
\rowcolor{lightgray}
\textbf{Ours}        & \textbf{29.6} & { \textbf{38.4}} & \textbf{45.0} & { \textbf{37.5}} & 47.0 & { \textbf{54.9}} & \textbf{42.8} & { \textbf{55.0}} & \textbf{56.0} & { \textbf{58.6}} & \textbf{75.4} & { \textbf{77.5}} \\
\bottomrule
\end{tabular}
\caption{Comparison on MLLMU, CLEAR, and UMU benchmarks. $\downarrow$ Lower is better for Forget VQA accuracy (forgetting of target visual knowledge), $\uparrow$ Higher is better for the others (retention of non-target visual knowledge and textual knowledge). For MLLMU and UMU, the Forget/Retain QA tasks are evaluated using the Accuracy, while for CLEAR, these tasks are evaluated using the ROUGE-L metric.}
\label{tab1}
\vspace{-1.5em} 
\end{table*}

\vspace{-1em} 
\section{Evaluation}
\vspace{-0.8em}
\subsection{Experiment setups}
\vspace{-0.8em}
Evaluating an unlearning method typically involves two stages. In the first stage, the \emph{base} model is trained to acquire a specific set of knowledge, comprising both a forgetting subset and a retained subset, resulting in a \emph{vanilla} model. In the second stage, the vanilla model is required to erase the forgetting subset, thereby producing an \emph{unlearned} model. This protocol is necessary because effective forgetting presupposes that the model has indeed learned the target knowledge. However, in practice, it is often unclear whether a base model has already internalized a particular piece of knowledge. Consequently, fictitious or synthetic profiles are commonly constructed as target knowledge to facilitate controlled and reliable evaluation of unlearning.

\textbf{Models and Datasets.} 
In our experiments, we employ LLaVA-1.5-7B-hf and Qwen2-VL-7B-Instruct as base models.
We perform evaluation on three benchmarks: MLLMU-Bench~\cite{liu2025protecting}, UMU-Bench~\cite{wang2025umu}, and CLEAR~\cite{dontsov2025clear}.
\textbf{Baseline Methods.}
To rigorously evaluate the performance of our approach, we benchmark it against a comprehensive suite of 7 baseline methods, ranging from foundational LLM unlearning techniques to state-of-the-art multimodal-specific methods:
\textbf{GA}~\cite{thudi2022unrolling} performs gradient ascent on the forget set.
\textbf{GA\_Diff}~\cite{liu2022continual} combines gradient ascent on the forget set with gradient descent on the retain set.
\textbf{KL\_Min}~\cite{nguyen2020variational} preserves utility by matching the vanilla model on the retain set via KL minimization while encouraging divergence on the forget set.
\textbf{NPO}~\cite{zhang2024negative} formulates unlearning as preference optimization with the forget set as negative samples and an oracle trained on the retain set.
\textbf{MANU}~\cite{liu2025modality} applies modality-aware neuron pruning to suppress forget-related behaviors.
\textbf{MMU(nlearner)}~\cite{huo2025mmunlearner} uses saliency-guided updates to target visual forgetting while protecting non-target parameters.

We adopt a suite of evaluation metrics to comprehensively assess both the forgetting and retention of visual and textual knowledge. Specifically, we use average accuracy on VQA tasks to quantify the extent to which visual knowledge is forgotten or preserved, and employ Text QA tasks to probe the retention of textual knowledge. In addition, we leverage open-ended generation tasks for evaluation, where ROUGE-L~\cite{lin2004rouge} is used to measure the textual overlap between the model’s generated outputs and the corresponding ground-truth responses.
All metrics are reported on the Forget Set, Retain Set, and Real-world Set, enabling a rigorous evaluation of unlearning effectiveness, retention fidelity, and the maintenance of general model utility. 

\vspace{-1em} 
\subsection{Main Results}
\paragraph{Effectiveness of Unlearning.}
Across the UMU-Bench, CLEAR, and MLLMU-Bench benchmarks, our method consistently achieves the strongest visual forgetting performance on both LLaVA-7B and Qwen2-VL, as shown in Table~\ref{tab1}. Specifically, with the LLaVA-7B model, our approach outperforms the state-of-the-art MMU by substantially reducing Forget VQA accuracy from $46.0\%$ to $44.2\%$ on UMU-Bench, from $36.2\%$ to $32.4\%$ on CLEAR, and from $31.2\%$ to $29.6\%$ on MLLMU-Bench, respectively. These results indicate that our approach can effectively erase visual knowledge associated with the target entities. 

Moreover, our method can effectively erase visual knowledge while preserving the model’s textual knowledge intact.
On the Retain QA and Forget QA tasks, which rely purely on textual knowledge, our approach maintains exactly the \emph{same} accuracy as the vanilla model, as shown in Table~\ref{tab1}. This highlights a clear advantage over competing methods. We also observe that some baselines even outperform the vanilla model under the ROUGE-L metric (e.g., GA\_Diff achieves 0.125 on Forget QA with Qwen2-VL on the CLEAR benchmark), which is likely attributable to the imprecision and known limitations of ROUGE-L as an evaluation measure for this setting.


Preserving non-target visual knowledge is widely regarded as the most challenging aspect of MLLM unlearning. As evidenced by the overall results, methods that aggressively suppress the Forget VQA accuracy typically incur substantial collateral degradation on Retain VQA. In contrast, our approach effectively mitigates this degradation and achieves a more favorable balance. 
Specifically, with the LLaVA-7B model, our approach outperforms MMU by significantly improving Retain VQA accuracy from $54.4\%$ to $74.4\%$ on UMU-Bench, from $46.6\%$ to $50.0\%$ on CLEAR, and from $44.2\%$ to $45.0\%$ on MLLMU-Bench, respectively. We further attribute this improvement to the complementary roles of CVF and NCU, as analyzed in the following sections.

\subsection{Ablation Study}

\paragraph{Contrastive Visual Forgetting.}
To identify the contribution of our CVF mechanism to visual knowledge forgetting, we compare several simplified variants: (i) replacing our Eq.(\ref{eq:push}) of $\mathcal{L}_{\mathrm{CVF}\text{-}push}$ with a conventional MSE loss $\mathcal{L}_{\mathrm{nMSE}}$, and (ii) without the anchor pulling Eq.(\ref{eq:pull}) of $\mathcal{L}_{\mathrm{CVF}\text{-}pull}$.

From Table~\ref{tab2}, we have following conclusions: (1) Compared with the conventional $\mathcal{L}_{\mathrm{nMSE}}$, our $\mathcal{L}_{\mathrm{CVF}\text{-}push}$ reduces Forget VQA from Acc=$40\%$ to Acc=$26.4\%$, indicating that contrastive learning is more effective at erasing target visual knowledge under intermediate-level supervision. (2) Without $\mathcal{L}_{\mathrm{CVF}\text{-}pull}$, the model suffers the most severe degradation on both Retain VQA ($31.5\%$ vs. $38.8\%$) and Real-world VQA ($40.6\%$ vs. $41.4\%$). This clearly indicates that the anchor pulling scheme plays a crucial role in mitigating IVRs drift and in preserving non-target visual knowledge.

\begin{wraptable}{r}{0.52\columnwidth}
\vspace{-0.8em}
\centering
\scriptsize
\setlength{\tabcolsep}{1pt}
\renewcommand{\arraystretch}{0.9}
\begin{tabular}{l|cc|cc|cc}
\toprule
\multirow{2}{*}{\textbf{Method / Variant}} & \multicolumn{2}{c|}{\textbf{Forget}} & \multicolumn{2}{c|}{\textbf{Retain}} & \multicolumn{2}{c}{\textbf{Real-world}} \\
& \textbf{VQA $\downarrow$} & \textbf{QA $\uparrow$} & \textbf{VQA $\uparrow$} & \textbf{QA $\uparrow$} & \textbf{VQA $\uparrow$} & \textbf{QA $\uparrow$} \\
\midrule
Vanilla & 45.8 & 38.4 & 45.2 & 37.5 & 47.4 & 54.9 \\
$\mathcal{L}_{\mathrm{nMSE}}$ & 40.0 & 33.6 & 44.3 & 31.5 & 46.6 & 53.6 \\
\midrule
$\mathcal{L}_{\text{CVF-push}}$ & 26.4 & 38.4 & 31.5 & 37.5 & 40.6 & 54.9 \\
\hspace{0.5em} + $\mathcal{L}_{\text{CVF-pull}}$ & 28.3 & 38.4 & 38.8 & 37.5 & 41.4 & 54.9 \\
\hspace{1.0em} + $\mathcal{L}_{\text{RET}}$ & 29.6 & 38.4 & 40.7 & 37.5 & 42.8 & 54.9 \\
\hspace{1.5em} + $\mathcal{L}_{\text{GUM}}$ (Full CVF) & 32.2 & 38.4 & 43.2 & 37.5 & 45.6 & 54.9 \\
\midrule
CVF + Random & 32.4 & 38.4 & 43.6 & 37.5 & 44.8 & 54.9 \\
\rowcolor{gray!15}
\textbf{CVF + NCU (Ours)} & \textbf{29.6} & \textbf{38.4} & \textbf{45.0} & \textbf{37.5} & \textbf{47.0} & \textbf{54.9} \\
\bottomrule
\end{tabular}
\caption{Ablation study. ``Random'' denotes random subspace unlearning.}
\label{tab2}
\vspace{-1.0em}
\end{wraptable}

Besides, we further investigate the contribution of knowledge retention Eq.(\ref{eq:ret}) of $\mathcal{L}_{\text{RET}}$ and the general utility maintenance Eq.(\ref{eq:gum}) of $\mathcal{L}_{\text{GUM}}$. Accordingly, we have following conclusions: (1) Without $\mathcal{L}_{\text{RET}}$, the model suffers a further degradation in Retain VQA ($38.8\%$ vs. $40.7\%$), demonstrating that explicit retain-set supervision is effective in balancing the forgetting–retention trade-off. (2) Finally, removing $\mathcal{L}_{\text{GUM}}$ leads to a noticeable degradation in general multimodal utility, with Real-world VQA ($42.8\%$ vs. $45.6\%$). This indicates that output-level supervision remains crucial for preserving overall multimodal capability.

\paragraph{Null-space–Constrained Unlearning.}
To identify the contribution of our NCU scheme in decoupling the forgetting operation from knowledge retention, we compare the following variants: (i) \emph{Full CVF-only}, which adopts standard LoRA-based unlearning without any null-space constraint;
(ii) \emph{CVF + Rand}, which replaces our NCU scheme with \emph{Random subspace unlearning}. Specifically, we replace our NCU basis with a randomly constructed orthonormal basis, representing a random subspace unrelated to retained knowledge. Concretely, for each layer $\ell$, we sample a random matrix $G_{\ell}\in R^{d_{\mathrm{in}}\times r}$ and perform QR factorization to obtain an orthonormal basis $Q_{\ell}\in R^{d\times r}$, then set $A_{\ell}=Q_{\ell}^{\top}$ and freeze it throughout unlearning while training only $B_{\ell}$. This $A_{\ell}$ is not associated with the retained knowledge and therefore does not explicitly preserve retention knowledge.

From Table~\ref{tab2}, we draw the following conclusions: (1) NCU consistently improves the forgetting–retention trade-off compared with the variant without the null-space constraint. Across all benchmarks, \emph{CVF + NCU} can reduce Forget VQA accuracy compared to the \emph{Full CVF-only} variant ($29.6\%$ vs. $32.2\%$), while delivering substantially stronger retention on Retain VQA ($45.0\%$ vs. $43.2\%$), and Real-world VQA ($47.0\%$ vs. $45.6\%$). (2) \emph{Random subspace unlearning} fails to achieve comparable gains. It yields poorer performance on Forget VQA ($32.4\%$ vs. $29.6\%$) and fails to better preserve accuracy on both Retain VQA ($43.6\%$ vs. $45.0\%$) and Real-world VQA ($44.8\%$ vs. $47.0\%$).
These results demonstrate that NCU not only integrates seamlessly with the CVF scheme but also synergistically amplifies the effectiveness of both components, leading to a superior balance between effective forgetting and knowledge retention.

\begin{figure*}[!t]
    \centering
    \includegraphics[width=1.05\linewidth]{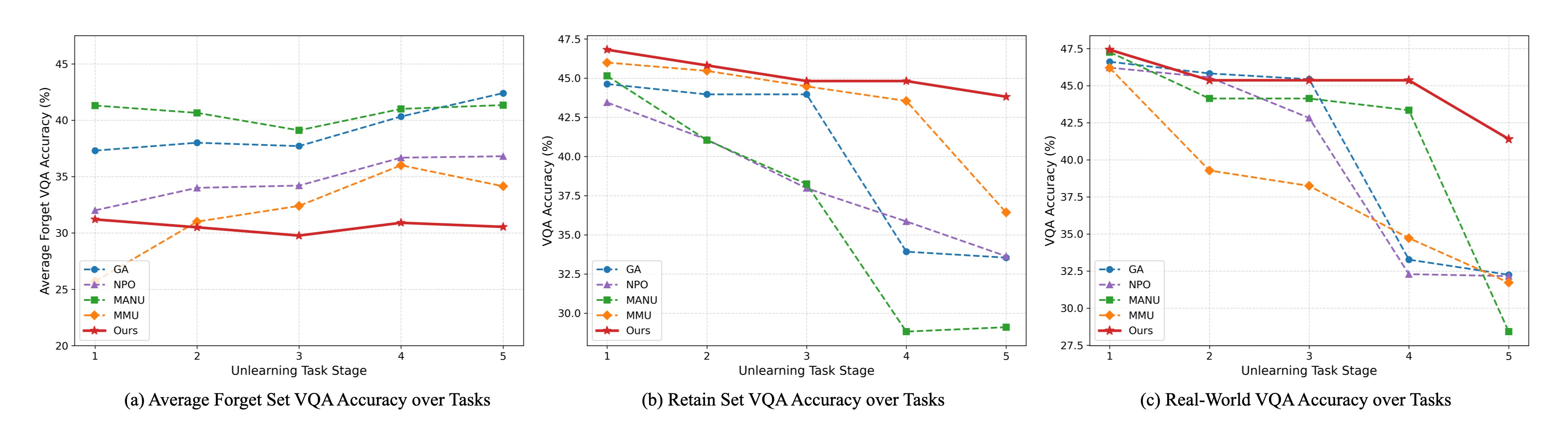}  
    \caption{Evaluation of continual unlearning with five sequential forgetting tasks. (a) Average VQA accuracy over cumulative Forget sets across tasks. (b) VQA accuracy on the Retain set across tasks. (c) VQA accuracy on the Real-world set across tasks.}    
    \label{fig2}
    \vspace{-1em}
\end{figure*}

\subsection{Continual Unlearning Evaluation.}

Continual unlearning is inherently more difficult than static unlearning, primarily due to the exacerbated forgetting–retention trade-off. Under sequential forgetting requests, the unlearning operation must be performed repeatedly, and each new forgetting operation may interfere with previously completed ones, leading to cumulative performance degradation over time. In contrast, our approach is explicitly designed to achieve a favorable balance between forgetting and retention, which substantially mitigates the accumulation of such degradation in continual unlearning.

To evaluate the performance under continual unlearning, we randomly split the Forget set of MLLMU-Bench into five disjoint subsets, thereby simulating five sequential unlearning tasks. Except for the Forget set, both the Retain set and the Real-world set remain identical to those used in the static unlearning setting.


The x-axis denotes the five forgetting tasks arriving sequentially, while the y-axis reports the corresponding accuracy on the Retain VQA and Real-world VQA tasks. As shown in Fig.~\ref{fig2} (b) and (c), as continual unlearning progresses, the VQA accuracy on retained knowledge gradually degrades. 
Notably, for the MANU method, the degradation on the Retain Set becomes substantially more severe starting from the 2nd task, whereas the degradation on the Real-world Set only becomes pronounced from the 5th task. In contrast, for the MMU method, the Real-world Set degrades much more severely beginning from the 2nd task, while the degradation on the Retain Set becomes evident only from the 5th task.
However, our approach exhibits a much slower degradation on both Retain and Real-world Sets than the competing methods, demonstrating its clear advantage in continual unlearning scenarios.

Fig.~\ref{fig2} (a) further illustrates the dynamics of accuracy on the forgetting knowledge. As new subsets are sequentially unlearned, the forgetting effectiveness on previously requested subsets is only \emph{mildly affected} across all methods. Moreover, our approach consistently outperforms the other methods throughout the entire process, demonstrating more stable and effective forgetting under continual unlearning.

Our approach demonstrates a clear advantage over existing unlearning methods. 
We attribute this gain to the complementary effects of CVF and NCU, which effectively improve the forgetting–retention trade-off at each unlearning step. As a result, the benefits accumulate over successive unlearning operations, leading to substantially improved performance in the continual unlearning setting.

\vspace{-0.8em}

\section{Conclusion}
\vspace{-0.8em}
In this paper, we present a novel MLLM unlearning approach that achieves a robust forgetting--retention trade-off while selectively separating target visual knowledge from retained knowledge.
Our CVF mechanism leverages intermediate-level supervision to guide the representations of target visual concepts toward appropriate regions in the feature space, while preventing degradation of retained knowledge.
Furthermore, we propose a null-space-constrained unlearning strategy that explicitly decouples forgetting operations from knowledge retention. These two components synergistically reinforce each other, enabling a superior balance between effective forgetting and robust knowledge retention.
Extensive experiments demonstrate that our method significantly outperforms existing approaches. In addition, we are the first to investigate the problem of continual MLLM unlearning, and our approach exhibits clear and consistent advantages in this more challenging setting.

\bibliographystyle{unsrtnat}
\bibliography{refs}


\newpage
\appendix
\onecolumn
\section{Motivation and Problem Formulation}

Recent literature presents two competing definitions of MLLM unlearning: (i) forgetting \emph{visual} knowledge while preserving the corresponding \emph{textual} knowledge, and (ii) forgetting both \emph{visual} and \emph{textual} knowledge of the target entity. These two definitions have been proposed in prior works such as MANU~\cite{liu2025modality}, MMU~\cite{huo2025mmunlearner}, and MMNeuron~\cite{huo2024mmneuron}. In this paper, we adopt the first definition because it explicitly highlights the need to disentangle visual and textual knowledge. Once the two modalities are properly decoupled, both definitions can be handled in a unified manner: for definition (i), we remove only the visual-side knowledge after disentanglement; for definition (ii), we can sequentially erase the visual and textual knowledge in a controlled way.


We attribute this phenomenon to the \emph{distributed} nature of visual knowledge representation in MLLMs. Visual semantics are not stored in a small set of isolated neurons; rather, they are encoded in a shared embedding subspace spanning multiple layers and feature channels. Consequently, na\"{i}ve structural pruning disrupts this distributed topology and can destabilize cross-modal alignment, leading to a collapse in multimodal representations.

\section{Related Work}
\textbf{Machine Unlearning for LLMs.}
Machine Unlearning (MU) has become an important paradigm for mitigating the memorization of sensitive, private, or copyrighted data in large language models. A representative line of work is based on parameter updates that directly suppress the generation of target knowledge. For example, Gradient Ascent (GA)~\cite{thudi2022unrolling} performs unlearning by maximizing the loss on the forget set, thereby discouraging the model from reproducing target content. A central difficulty in this line of research is the forgetting--utility trade-off, since aggressive forgetting often harms the model's performance on unrelated knowledge and downstream tasks. To alleviate this issue, subsequent methods introduce regularization or preservation constraints, such as KL-divergence matching~\cite{maini2024tofu} and retain-set supervision~\cite{rafailov2023direct}. Reinforcement-learning-based formulations have also been explored. For instance, Negative Preference Optimization (NPO)~\cite{zhang2024negative} treats forget samples as negative preferences and aims to improve unlearning stability through preference optimization.

Recent studies have further examined the optimization and efficiency aspects of LLM unlearning. GRU~\cite{wang2025gru} analyzes the conflict between forget and retain objectives from the perspective of gradient geometry, and reduces this conflict by projecting unlearning gradients away from retain-related directions. In contrast to parameter-update-based methods, ERASE~\cite{muresanufast} reformulates unlearning as an in-context inference procedure, using retrieval and example selection instead of explicit retraining. These works collectively show that LLM unlearning has evolved from simple forgetting objectives toward more structured and efficient mechanisms for balancing forgetting and preservation.

\textbf{Machine Unlearning for MLLMs.}
Machine unlearning for multimodal large language models is still at an early stage. Existing studies suggest that MLLMs introduce additional complexity beyond text-only LLMs because visual and textual information interact within a shared multimodal model~\cite{liu2024large,wang2025mllm,cohen2025performance,yu2024understanding}. As a result, unlearning in MLLMs is not yet defined in a fully unified way. One line of work formulates MLLM unlearning as removing target \emph{visual} knowledge while preserving textual knowledge and general utility~\cite{wang2025mllm,huo2025mmunlearner}. Another line adopts a broader forgetting objective and aims to suppress both visual and textual knowledge associated with the target entity~\cite{liu2025modality}.

Existing MLLM unlearning methods also differ substantially in their technical designs. SIU~\cite{li2024single} studies the removal of visual patterns associated with real-world entities through multi-faceted fine-tuning. MMUnlearner~\cite{huo2025mmunlearner} focuses on visual forgetting while preserving non-target behavior through multimodal-specific optimization. MANU~\cite{liu2025modality} performs modality-aware neuron pruning to suppress target-related knowledge more structurally. VKD~\cite{wang2025mllm} introduces a visual-knowledge-distillation strategy, where a teacher model provides visual-layer constraints to reduce collateral damage during unlearning. Although these methods provide important first steps, the core challenge of MLLM unlearning remains unresolved: how to erase target visual knowledge while preserving retained knowledge and general multimodal utility in a selective and stable manner.

Our work is most closely related to methods that study selective visual unlearning in MLLMs. Compared with prior approaches, we focus on structured forgetting at the intermediate visual representation level and explicitly constrain update directions to reduce interference with retained knowledge.

\section{Implementation Details}
\subsection{Datasets}
Given the uncertainty of whether a pretrained MLLM already internalizes the target knowledge in a benchmark, we first fine-tune each backbone on the full training split of the benchmark to obtain a \emph{vanilla} model that explicitly memorizes the target entities. We then apply different unlearning algorithms to this vanilla model and evaluate forgetting effectiveness and utility preservation under the same evaluation protocol.

We conduct experiments on three representative multimodal unlearning benchmarks: MLLMU-Bench~\cite{liu2025protecting}, UMU-Bench~\cite{wang2025umu}, and CLEAR~\cite{dontsov2025clear}. All benchmarks follow a similar forget/retain/real-world evaluation structure: the Forget Set contains samples associated with target entities to be removed, the Retain Set contains non-target entities to measure preservation of remaining knowledge, and the Real-world Set consists of out-of-distribution profiles to assess general multimodal utility beyond both forget and retain data.

\paragraph{MLLMU-Bench~\cite{liu2025protecting}.}
MLLMU-Bench is a foundational benchmark designed for multimodal entity unlearning. It includes 500 fictitious profiles and 153 public celebrity profiles. Each profile is paired with portrait images and a set of 14 customized question--answer pairs spanning both multimodal and unimodal settings. The benchmark explicitly supports probing visual knowledge and textual knowledge separately: multimodal questions require correct visual recognition grounded in the input image, whereas text-only questions probe whether textual facts are preserved when the image is omitted. Following the official protocol, we evaluate unlearning on the Forget Set, preservation on the Retain Set (non-target fictitious profiles), and general utility on the Real-world Set (celebrity profiles), which is disjoint from the targeted entities.

\paragraph{UMU-Bench~\cite{wang2025umu}.}
UMU-Bench extends the MLLMU-Bench setting with a stronger focus on modality consistency and alignment. It contains 653 individual profiles, each described through carefully curated unimodal and multimodal knowledge. In contrast to benchmarks where unlearning can be partially ``hidden'' by modality shortcuts, UMU-Bench is designed to reveal cases where knowledge appears removed in one modality but still persists in another. Concretely, it provides paired probes that test whether forgetting on visual queries is consistent with the model's behavior on text-only queries, thereby offering a more stringent evaluation of whether an unlearning method truly removes the intended visual concept while maintaining stable textual behavior and preserving non-target multimodal capabilities.

\paragraph{CLEAR~\cite{dontsov2025clear}.}
CLEAR extends the text-only TOFU benchmark into a multimodal unlearning setting. For each author profile in TOFU, CLEAR augments the data with corresponding face images and detailed textual descriptions generated by GPT-4o, enabling controlled evaluation of visual and textual unlearning for identity-related knowledge. Similar to MLLMU-Bench, CLEAR is organized into Forget, Retain, and Real-world subsets, which allows us to quantify: (i) the degree to which targeted visual knowledge is removed, (ii) how well non-target knowledge is preserved, and (iii) whether the model retains general utility on real-world evaluation that is unrelated to both forget and retain entities. This benchmark is particularly useful for assessing how visual augmentation changes unlearning dynamics compared to purely text-based unlearning settings.


\paragraph{Continual Unlearning Task Dataset}
To evaluate continual multimodal unlearning, we construct a sequential unlearning setting based on MLLMU-Bench~\cite{liu2025protecting}.
We randomly partition the original \textbf{Forget Set} (15\% of profiles) into five disjoint subsets, forming five sequential unlearning tasks $\{T_1,\dots,T_5\}$ that simulate five consecutive forgetting requests.
Each task contains 15 target profiles (i.e., 15 forgetting units) and represents a distinct visual entity knowledge to be removed.
Across all tasks, we keep the \textbf{Retain Set} (the remaining 85\% of profiles) and the \textbf{Real-world Set} identical to those used in the static unlearning setting, so that continual results are directly comparable.

The model performs unlearning sequentially from $T_1$ to $T_5$.
After completing task $T_t$, we evaluate (i) \textbf{Forget VQA} and \textbf{Forget QA} on the currently unlearned task $T_t$ to measure immediate forgetting effectiveness, and (ii) the same Forget metrics on all previously unlearned tasks $\{T_1,\dots,T_{t-1}\}$ to quantify whether earlier forgetting effects are preserved or overwritten over time.
In parallel, we report \textbf{Retain VQA} and \textbf{Retain QA} on the shared Retain Set to measure knowledge preservation, and evaluate general utility on the shared Real-world Set throughout the entire sequence.

\begin{figure*}[htbp!]
    \centering
    \includegraphics[width=1\linewidth]{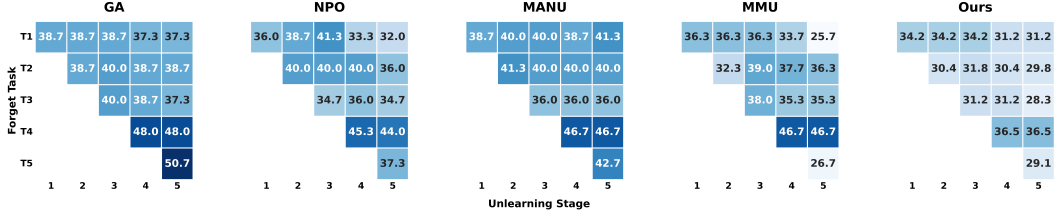}  
    \caption{Stage-wise performance heatmaps of different unlearning methods. The color intensity represents VQA accuracy, where lighter colors indicate lower accuracy (better unlearning efficacy).}    
    \label{figHeat}
\end{figure*}

To intuitively visualize the efficacy of our proposed framework in the continuous unlearning setting, we present the stage-wise performance heatmaps for all compared methods in Figure \ref{figHeat}. The horizontal axis represents the sequential unlearning stages (1 to 5), while the vertical axis corresponds to the specific tasks evaluated ($T_1$ to $T_5$).

\subsection{Evaluation Metrics}
We follow the evaluation protocols of each benchmark and report results on the Forget, Retain, and Real-world splits.
For multiple-choice VQA/QA questions, we use average accuracy as the primary metric.
For open-ended generation tasks, we adopt ROUGE-L~\cite{lin2004rouge}, which measures the similarity between the generated response and the reference answer based on the Longest Common Subsequence (LCS).
Let \(L_G\) and \(L_P\) denote the lengths of the reference and generated texts, respectively. Recall and precision are defined as:

\begin{equation}
\text{Recall} = \frac{\text{LCS}}{L_G}, \quad 
\text{Precision} = \frac{\text{LCS}}{L_P}.
\end{equation}

The final ROUGE-L score is computed as their harmonic mean:

\begin{equation}
\text{ROUGE-L} = 2 \cdot \frac{\text{Recall} \cdot \text{Precision}}{\text{Recall} + \text{Precision}}.
\end{equation}

ROUGE-L rewards both content coverage and faithful phrasing, and is widely used to evaluate whether the model preserves the key information and structure required by the ground-truth answer in free-form generation.

\subsection{Hyperparameter settings}
In multimodal unlearning evaluation, we first ensure that the backbone model has a clear and stable memory of the target entities, and then perform unlearning to assess the forgetting--retention trade-off. Therefore, for MLLMU-Bench~\cite{liu2025protecting} and CLEAR~\cite{dontsov2025clear}, we follow the configurations in the original papers and their public implementations to construct the vanilla model, and run all unlearning algorithms on top of it. In addition, for multimodal-specific baselines on these benchmarks (e.g., MANU~\cite{liu2025modality} and MMU~\cite{huo2025mmunlearner}), their papers and codebases typically provide well-validated hyperparameter settings, covering forgetting strength, training schedule, learning rate, regularization weights, and other optimization details. To ensure reproducibility and fair comparison, we directly adopt these established configurations on MLLMU-Bench and CLEAR.

In contrast, UMU-Bench places greater emphasis on cross-modal consistency and robustness of modality alignment, making different methods more sensitive to the forgetting strength and retention constraints. To avoid unnecessary bias caused by directly reusing default settings from other benchmarks, we re-organize and explicitly report the key hyperparameters used in the unlearning stage on UMU-Bench for both our method and the compared baselines. Concretely, we align the epoch budget, batch size, optimizer, and learning rate across methods, ensuring that all approaches are compared under the same computational budget and training protocol. The full hyperparameter settings used to reproduce the unlearning results on UMU-Bench are summarized in Table~\ref{tab3}.
In addition, we report the hyperparameter settings used for the vanilla memorization fine-tuning of the Qwen2-VL-7B model, as such fine-tuning details are not specified in the corresponding dataset papers.


\begin{table}[htbp!]
\centering
\resizebox{0.5\linewidth}{!}{
\begin{tabular}{lccccc}
\toprule
\textbf{Stage} & \textbf{Benchmark} & \textbf{Epochs} & \textbf{Batch Size} & \textbf{Learning Rate} \\
\midrule
\multirow{3}{*}{Vanilla} 
 & UMU-Bench   & 4 & 8  & $6 \times 10^{-4}$ \\
 & MLLMU-Bench & 5 & 8  & $4 \times 10^{-4}$ \\
 & CLEAR       & 5 & 8  & $3 \times 10^{-4}$ \\
\midrule
\multirow{7}{*}{Unlearning} 
 & GA            & 3 & 4  & $5 \times 10^{-6}$ \\
 & GA\_Diff      & 3 & 4  & $5 \times 10^{-6}$ \\
 & KL\_Min       & 3 & 4  & $5 \times 10^{-6}$ \\
 & NPO           & 3 & 4  & $5 \times 10^{-6}$ \\
 & MANU          & 4 & 4  & $5 \times 10^{-6}$ \\
 & MMU   & 4 & 4  & $5 \times 10^{-6}$ \\
 & \textbf{Ours} & 4 & 4  & $5 \times 10^{-6}$ \\
\bottomrule
\end{tabular}
}
\caption{Hyperparameter settings for the vanilla memorization stage and the subsequent unlearning stage on UMU-Bench and related benchmarks.}
\label{tab3}
\vspace{-1.8em} 
\end{table}

\paragraph{Hyperparameter selection and sensitivity.}
Our method involves two levels of weighting coefficients. 
At the inner level, $\lambda$ balances the \emph{push} and \emph{pull} terms inside the CVF objective,
\begin{equation}
\mathcal{L}_{\mathrm{CVF}}
=\lambda\,\mathcal{L}_{\mathrm{CVF\text{-}push}}+\mathcal{L}_{\mathrm{CVF\text{-}pull}},
\end{equation}
which mainly controls the \emph{directionality} of representation movement (repulsion from the reference versus attraction to the retain-domain anchor). 
At the outer level, $\alpha$ and $\beta$ weight the three loss components,
\begin{equation}
\min_{\theta}\;
\alpha\,\mathcal{L}_{\mathrm{CVF}}
+\beta\,\mathcal{L}_{\mathrm{RET}}
+\mathcal{L}_{\mathrm{GUM}},
\end{equation}
thereby adjusting the global trade-off among target forgetting, non-target retention, and general multimodal utility.

\begin{figure*}[htbp!]
    \centering
    \includegraphics[width=1\linewidth]{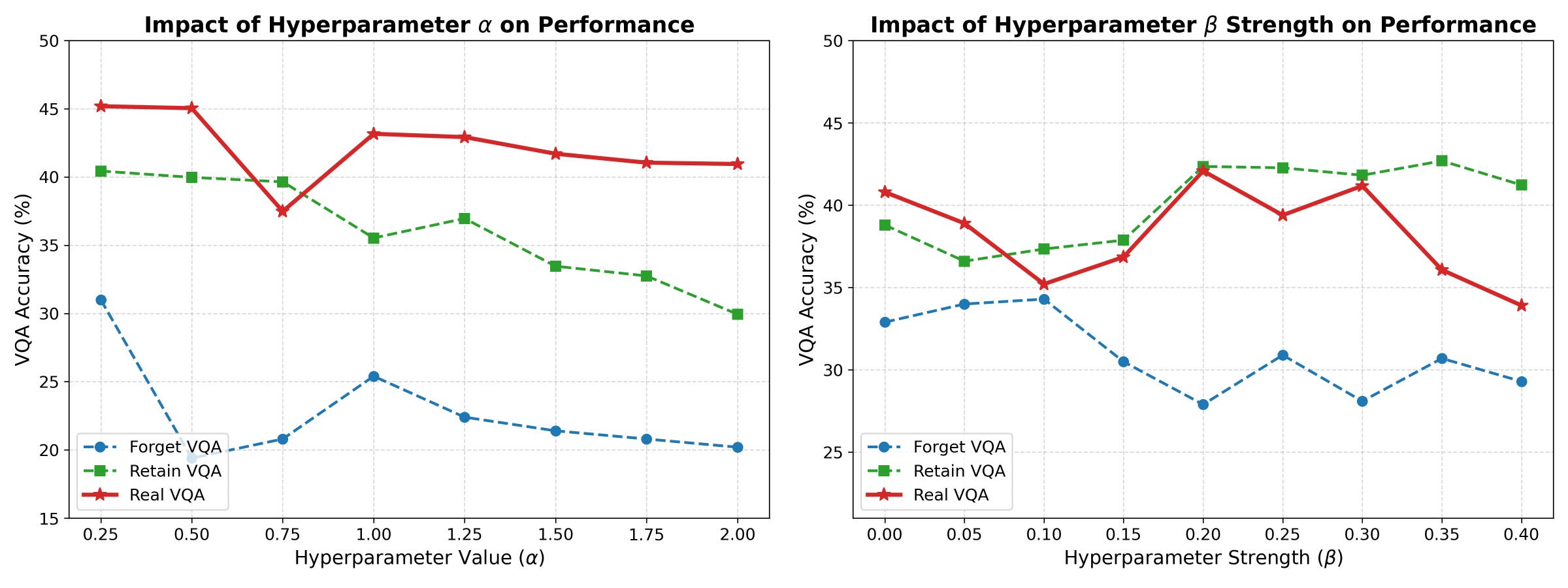}  
    \caption{Hyperparameter sensitivity of $\alpha$ and $\beta$. Performance trends on Forget/Retain/Real-World VQA when varying $\alpha$ and $\beta$ while keeping other settings fixed.}    
    \label{fig3}
\end{figure*}

Since $\lambda$ is an \emph{inner-level} balancing factor that primarily affects the stability of CVF, we select $\lambda$ via an automated validation-based tuning procedure. 
Concretely, we perform a lightweight search over a small candidate set and choose the value that best maintains retention and general utility while achieving the desired forgetting strength on a held-out validation split.\footnote{We use the same training budget for each candidate and do not reuse test sets for selection.}

In contrast, $\alpha$ and $\beta$ operate at the \emph{outer level} and directly govern the overall objective composition, which can affect the full forgetting--retention--utility trade-off. 
We therefore conduct explicit sensitivity experiments for $\alpha$ and $\beta$, reporting performance trends on Forget/Retain/Real-World VQA in Fig.~\ref{fig3}.

\section{Additional Experiments}
\subsection{Experimental Results on MLLMU-Bench}
We further evaluate all methods under stronger unlearning settings by increasing the forget ratio to 10\% and 15\% on MLLMU-Bench, in order to examine robustness against varying unlearning intensity.
As reported in Table~\ref{tab4} and Table~\ref{tab5}, our approach remains stable across both ratios.
When forgetting 10\% or 15\% of the target data, our method consistently achieves substantially lower accuracy on Forget VQA than competing baselines, indicating a stronger ability to erase the target visual concepts.
Meanwhile, it maintains high performance on Retain VQA and Real-world QA, suggesting that the forgetting process introduces only minimal collateral damage and preserves general utility.
Overall, these results demonstrate that our approach enables stable and controllable unlearning without sacrificing textual understanding or multimodal alignment: it effectively removes target visual knowledge while keeping text semantics and non-target capabilities largely intact, highlighting strong generalization across different forgetting strengths.

\begin{table}[htbp!]
\centering
\scriptsize  
\setlength{\tabcolsep}{2pt} 
\renewcommand{\arraystretch}{0.8} 

\begin{tabular}{l|cccccc}
\toprule
\multirow{3}{*}{\textbf{Methods}} &
\multicolumn{6}{c}{\textbf{10\% Forget ratio}} \\

\cmidrule(lr){2-7}

& Forget VQA & Forget QA & Retain VQA & Retain QA & RW VQA & RW QA\\
& Acc($\downarrow$) & Acc($\uparrow$) & Acc($\uparrow$) & Acc($\uparrow$) & Acc($\uparrow$) & Acc($\uparrow$)\\
\midrule
GA          & 41.7 & 45.4 & 47.5 & 33.2 & 32.2 & 54.6 \\
GA\_Diff    & 39.3 & 43.4 & 46.5 & 41.2 & 39.3 & 53.5 \\
KL\_Min     & 42.5 & 45.3 & 47.0 & 32.0 & 32.1 & 54.9\\
NPO         & 41.7 & 44.6 & 47.4 & 32.7 & 32.2 & 54.9 \\
MMU         & 32.2 & 44.2 & 46.5 & 43.2 & 39.0 & 54.2 \\

\rowcolor{lightgray}
\textbf{Ours}       & 31.6 & 46.3 & 47.5 & 45.2 & 39.8 & 54.9\\
\bottomrule
\end{tabular}

\caption{
Performance comparison of different methods at 10\% training ratios across Forget, Retain, and Real-world tasks.}
\label{tab4}
\end{table}

\begin{table}[htbp!]
\centering
\scriptsize  
\setlength{\tabcolsep}{2pt} 
\renewcommand{\arraystretch}{0.8} 

\begin{tabular}{l|cccccc}
\toprule
\multirow{3}{*}{\textbf{Methods}} &
\multicolumn{6}{c}{\textbf{15\% Forget ratio}} \\
\cmidrule(lr){2-7}

& Forget VQA & Forget QA & Retain VQA & Retain QA & RW VQA & RW QA \\
& Acc($\downarrow$) & Acc($\uparrow$) & Acc($\uparrow$) & Acc($\uparrow$) & Acc($\uparrow$) & Acc($\uparrow$)\\

\midrule
GA          & 40.1 & 45.1 & 47.9 & 34.7 & 32.2 & 55.0 \\
GA\_Diff    & 38.5 & 40.6 & 45.8 & 34.1 & 31.0 & 54.0 \\
KL\_Min     & 40.3 & 44.6 & 47.0 & 34.1 & 31.8 & 54.8 \\
NPO         & 39.6 & 43.9 & 47.8 & 34.6 & 32.1 & 54.9 \\
MMU         & 29.3 & 43.2 & 46.1 & 38.4 & 31.3 & 53.9 \\

\rowcolor{lightgray}
\textbf{Ours}        & 29.8 & 45.1 & 47.9 & 38.6 & 33.2 & 55.0 \\

\bottomrule
\end{tabular}

\caption{
Performance comparison of different methods at 15\% training ratios across Forget, Retain, and Real-world tasks.}
\label{tab5}
\end{table}

\subsection{Performance of Baseline Methods under Visual Module Only Constraints}
Furthermore, to ensure a more rigorous and fair comparison with our proposed architecture, we adapted the baseline methods—including GA, GA\_Diff, MANU, and MMUnlearner—to operate under the same constraints. Specifically, we restricted their parameter update scope to align with ours: freezing the LLM backbone parameters and performing gradient updates exclusively on the Vision Encoder and Projector. The results are summarized in Table~\ref{tabvision}.

\begin{table}[htbp!]
\centering
\scriptsize  
\setlength{\tabcolsep}{2.0pt} 
\renewcommand{\arraystretch}{0.85} 
\begin{tabular}{l|cc|cc|cc}
\toprule
\multirow{3}{*}{\textbf{Methods}} &
\multicolumn{2}{c|}{\textbf{Forget Set}} &
\multicolumn{2}{c|}{\textbf{Retain Set}} &
\multicolumn{2}{c}{\textbf{Real-world Set}} \\
& Forget VQA & Forget QA & Retain VQA & Retain QA & RW VQA & RW QA \\
& Acc($\downarrow$) & Acc($\uparrow$) & Acc($\uparrow$) & Acc($\uparrow$) & Acc($\uparrow$) & Acc($\uparrow$) \\
\midrule
Vanilla                     & 45.8 & 38.4 & 45.2 & 37.5 & 47.4 & 54.9 \\
GA                       & 21.6 & 38.4 & 22.8 & 37.5 & 21.2 & 54.9\\
GA\_Diff                  & 16.4 & 38.4 & 22.5 & 37.5 & 22.8 & 54.9 \\

MMUNLEARNER              & 33.6 & 38.4 & 42.4 & 37.5 & 38.3 & 54.9 \\

\rowcolor{lightgray}
Ours & 29.6 & 38.4 & 45.0 & 37.5 & 47.0 & 54.9 \\
\bottomrule
\end{tabular}
\caption{Quantitative comparison of baselines and our approach when restricting the update scope to the \textbf{visual module only}.}
\label{tabvision}
\vspace{-1.8em} 
\end{table}

As shown in Table~\ref{tabvision}, when these baseline methods are confined to the visual module, approaches such as GA and GA\_Diff indeed exhibit a decrease in Forget VQA accuracy (indicating facilitated unlearning). However, this comes at the cost of a substantial decline in accuracy on both Retain VQA and Real-world VQA tasks. We attribute this to the fact that concentrating gradient updates solely on the front-end visual layers leads to the rapid destruction of the general feature representation capabilities for images.

\subsection{Efficiency of Unlearning.}
Our approach is computationally efficient because the unlearning stage freezes the LLM and only performs lightweight adaptation on the visual module.
We compare the average unlearning time per epoch across different methods on a server equipped with 4$\times$NVIDIA RTX 4090 GPUs, and report the results in Table~\ref{tab7}.
Overall, our approach achieves a favorable balance between unlearning effectiveness and computational efficiency.

This efficiency stems from two practical design choices.
First, the reference model does not require an additional model instance: it is realized by the same checkpoint with adapters disabled, so reference IVRs can be obtained without extra model loading overhead.
Second, the null-space constraint is computed only once via an offline calibration pass over the retain set to extract the layer-wise bases, and is reused throughout training.
Therefore, NCU introduces negligible runtime overhead during unlearning, while providing consistent gains in retention and general utility.

\begin{table}[htbp!]
\centering
\scriptsize
\setlength{\tabcolsep}{1.2pt}
\renewcommand{\arraystretch}{0.9}
\begin{tabular}{lcccc}
\toprule
\textbf{Methods} & \textbf{Fine-tuning Scope} & \textbf{Running Time} & \textbf{GPU memory}\\
\midrule
GA           & Full Model (Vision + LLM)   & 43.3~($\pm$1.5)min & 48.7~($\pm$0.38)GB \\
GA\_Diff      & Full Model (Vision + LLM)   & 57.4~($\pm$1.5)min & 50.1~($\pm$0.37)GB  \\
KL\_Min       & Full Model (Vision + LLM)   & 54.8~($\pm$1.5)min & 50.5~($\pm$0.38)GB \\
NPO           & Full Model (Vision + LLM)   & 61.3~($\pm$1.5)min & 68.3~($\pm$0.38)GB  \\
MMU          & Full Model (Vision + LLM)    & 23.6~($\pm$1.5)min & 50.5~($\pm$0.38)GB\\
\midrule
\rowcolor{lightgray}
Ours       & Vision    & \textbf{22.4~($\pm$1.3)min} & 48.7~($\pm$0.34)GB\\
\bottomrule
\end{tabular}
\caption{Comparison of Unlearning Efficiency. Average running time per epoch (in minutes) on NVIDIA RTX 4090 GPU.}
\label{tab7}
\vspace{-2.5em}
\end{table}

\subsection{Additional Comparisons and Update-Scope Analysis}
\label{app:comparisons_scope}

We provide additional comparisons to recent and closely related methods, together with an analysis of the update scope used in our method.

\paragraph{Additional comparison with SIU.}
We additionally adapt SIU as a supplementary baseline under our protocol. SIU was originally designed for single-image concept unlearning under MMUBench, whereas our setting studies entity-level visual unlearning with explicit preservation of textual knowledge under the Forget / Retain / Real-world evaluation protocol. We therefore report SIU as an adapted baseline rather than a fully matched one. Under the MLLMU-Bench 5\% forgetting setting, SIU-adapted shows a weaker forgetting-retention trade-off than our method. Specifically, our method achieves lower Forget VQA and higher Retain VQA / Retain QA under the matched evaluation protocol. The corresponding quantitative results are reported in Table~\ref{tab:siu_adapted}.

\begin{table}[htp!]
\centering
\scriptsize
\setlength{\tabcolsep}{4pt}
\renewcommand{\arraystretch}{0.95}
\begin{tabular}{l|cc|cc|cc}
\toprule
\multirow{2}{*}{Method} & \multicolumn{2}{c|}{Forget} & \multicolumn{2}{c|}{Retain} & \multicolumn{2}{c}{Real-world} \\
 & F-VQA $\downarrow$ & F-QA $\uparrow$ & R-VQA $\uparrow$ & R-QA $\uparrow$ & RW-VQA $\uparrow$ & RW-QA $\uparrow$ \\
\midrule
Ours & 29.6 & 38.4 & 45.0 & 37.5 & 47.0 & 54.9 \\
SIU-adapted & 42.4 & 33.6 & 43.8 & 32.0 & 46.6 & 53.6 \\
\bottomrule
\end{tabular}
\caption{Additional comparison with SIU adapted to our protocol on MLLMU-Bench with a 5\% forgetting ratio.}
\label{tab:siu_adapted}
\end{table}

\paragraph{Discussion of other related methods.}
Methods such as SAUCE~\cite{geng2025sauce} or SalUn~\cite{fan2023salun} are also related, but they are not directly matched plug-in baselines under our current task formulation. SAUCE assumes an SAE-based feature intervention pipeline and a substantially different concept-level evaluation setting, while SalUn was originally proposed for image classification/generation rather than selective entity-level visual forgetting in MLLMs with textual knowledge preservation. We therefore discuss these methods as related work, but do not treat them as directly comparable baselines under the same protocol.

\paragraph{Update-scope analysis.}
We further analyze the effect of the update scope. Besides our visual-side LoRA setting, we compare against projector-only tuning and a broader-update variant with the LLM unfrozen. These variants clarify why we freeze the LLM and perform unlearning mainly on the visual side. The results are shown in Table~\ref{tab:update_scope}. Projector-only tuning yields weaker forgetting, suggesting that the projector alone does not provide sufficient capacity for effective visual unlearning. In contrast, unfreezing the LLM introduces noticeably larger collateral degradation on retained textual knowledge and general utility. These results support our design choice of freezing the LLM and applying unlearning mainly to the visual module.

\begin{table}[htp!]
\centering
\scriptsize
\setlength{\tabcolsep}{4pt}
\renewcommand{\arraystretch}{0.95}
\begin{tabular}{l|cc|cc|cc}
\toprule
\multirow{2}{*}{Variant} & \multicolumn{2}{c|}{Forget} & \multicolumn{2}{c|}{Retain} & \multicolumn{2}{c}{Real-world} \\
 & F-VQA $\downarrow$ & F-QA $\uparrow$ & R-VQA $\uparrow$ & R-QA $\uparrow$ & RW-VQA $\uparrow$ & RW-QA $\uparrow$ \\
\midrule
Projector-only & 34.8 & 38.4 & 42.1 & 37.5 & 45.5 & 54.9 \\
LLM unfrozen & 28.7 & 34.9 & 41.9 & 34.1 & 44.6 & 52.8 \\
Ours & 29.6 & 38.4 & 45.0 & 37.5 & 47.0 & 54.9 \\
\bottomrule
\end{tabular}
\caption{Additional analysis of the update scope on MLLMU-Bench with a 5\% forgetting ratio.}
\label{tab:update_scope}
\end{table}

\newpage
\section{Case Study}

We provide qualitative examples to further illustrate the forgetting and retention behaviors of different methods.
Figure~4 compares the responses generated by our approach and three representative baselines on four tasks: \emph{Forget VQA}, \emph{Forget QA}, \emph{Retain VQA}, and \emph{Retain QA}.
Panels (a)--(d) correspond to GA, MANU, MMU, and our approach, respectively.

Consistent with the quantitative results in the main paper, the baselines exhibit limited capability in erasing target visual knowledge, and their outputs often contain grammatical errors or semantic inconsistencies.
In some cases, the unlearning process also harms non-target visual understanding, leading to degraded performance on retention-related queries.
In contrast, our approach achieves more selective forgetting of the target visual concept while maintaining fluent syntax and coherent semantics.
In particular, on \emph{Retain VQA}, it produces contextually appropriate and consistent responses, indicating that it better balances forgetting and retention while preserving multimodal reasoning ability.

\begin{figure}[htbp!]
    \centering
    \includegraphics[width=1\linewidth]{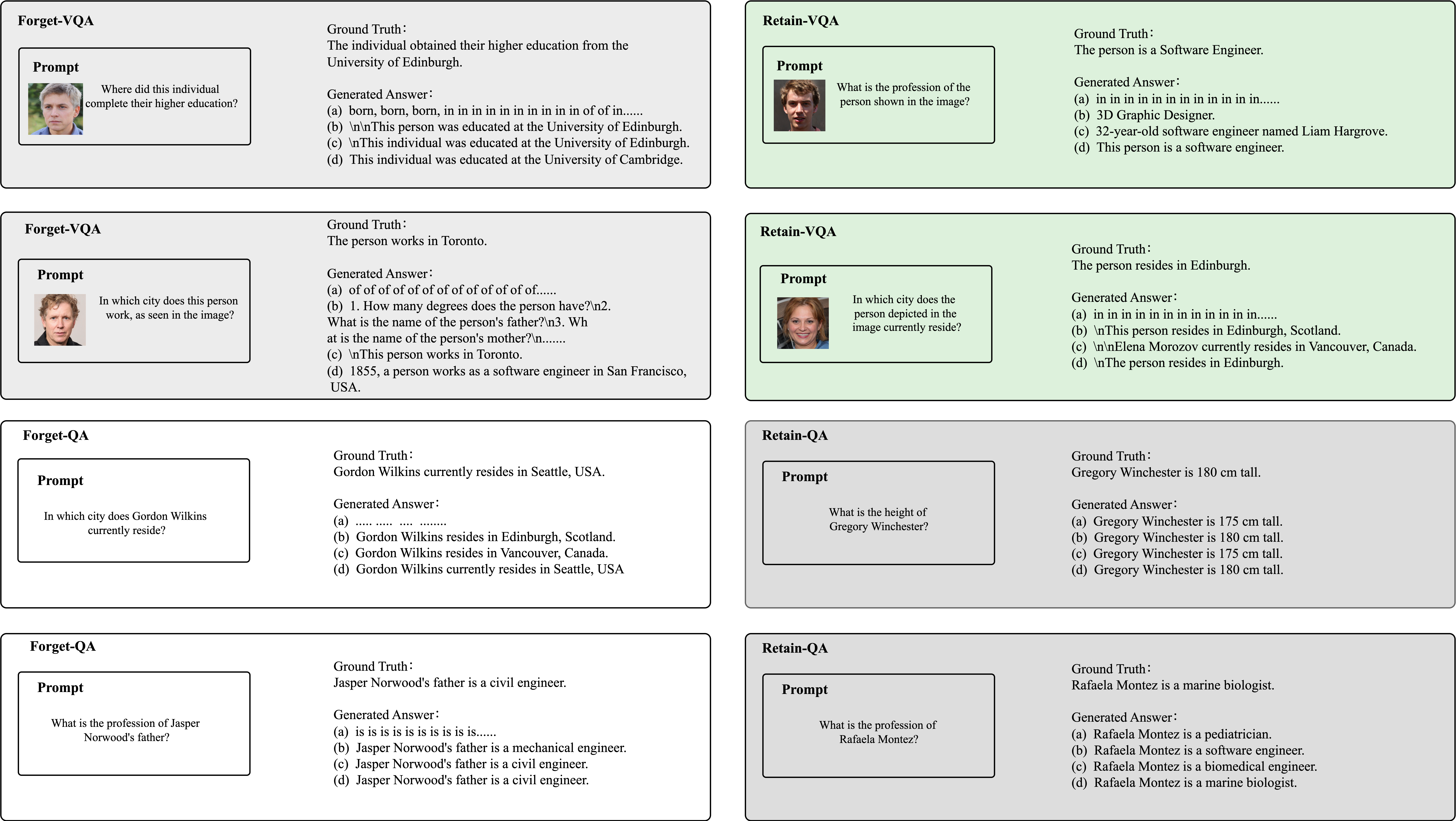}
    \caption{Qualitative results of GA, MANU, MMUNLEARNER, and our method on Forget/Retain VQA and Forget/Retain QA tasks.}
    \label{Sfig2}
\end{figure}


\end{document}